\documentclass[10pt]{article}
\usepackage[T1]{fontenc}
\usepackage{times}
\usepackage[letterpaper,textwidth=5.5in,textheight=9in,top=1in]{geometry}
\usepackage{graphicx}
\usepackage{booktabs}
\usepackage{float}
\usepackage{array}
\usepackage{caption}
\usepackage[hidelinks]{hyperref}
\usepackage{url}
\title{Are Causal Neuron Circuits Unique and Transferable?}
\author{\textbf{\"Omer Tekin Yavuz} \\ R\&D and AI Engineer \\ \.Ileti\c{s}im Yaz{\i}l{\i}m \\ Bursa, Turkey \\ \texttt{omtekyav@gmail.com}}
\date{}

\makeatletter
\renewcommand{\section}{\@startsection{section}{1}{\z@}{-11pt plus -2pt}{5pt}{\normalfont\large\bfseries}}
\renewcommand{\subsection}{\@startsection{subsection}{2}{\z@}{-8pt plus -2pt}{4pt}{\normalfont\normalsize\bfseries}}
\renewenvironment{abstract}{%
  {\centering\large\bfseries Abstract\par}\vspace{2pt}%
  \list{}{\leftmargin=0.5in\rightmargin=0.5in\listparindent=0pt\itemindent=0pt\topsep=0pt\partopsep=0pt\parsep=0pt}\item\relax}%
  {\endlist\vspace{4pt}}
\renewcommand{\@maketitle}{%
  {\centering
  \rule{\linewidth}{4pt}\par\vspace{8pt}
  {\LARGE\bfseries \@title\par}\vspace{6pt}
  \rule{\linewidth}{1pt}\par\vspace{16pt}
  {\setlength{\tabcolsep}{0pt}\renewcommand{\arraystretch}{1.05}\begin{tabular}[t]{c}\@author\end{tabular}\par}}%
  \vspace{14pt}}
\makeatother

\begin{document}
\maketitle

\begin{abstract}
Interpretability demonstrations often show that intervening on a group of neurons changes a model's answer. A recent example removes an 18-neuron group in Qwen3-32B and the model answers ``Simba'' instead of ``Scar'' to a two-hop Lion King riddle. Such a demonstration establishes that the group matters, but not how much of it is necessary, whether other groups could do the same, or whether the effect carries over to other inputs. We ask: when intervening on a group of neurons changes a model's answer, how much of that group is actually necessary? For every selected group we measure all 4,096 subsets of 12 MLP neurons. On a two-alternative capital task (Qwen3-0.6B), zero-ablating 12 selected neurons lowers the answer margin by 27\% while random neurons barely change it, yet replacing the same neurons with their values from other questions also barely changes it: their activations are nearly constant across questions, and we find no evidence that they carry question-specific content. On two-hop relation chains, patching 12 neurons with activations from the first-hop question shifts the answer to the bridge entity (from Scar to Simba): the greedy answer starts with the bridge in 9/11 chains for Qwen3-1.7B, 9/11 for Llama-3.2-1B and 8/10 for Gemma-3-1B; random neurons never do. The full group is rarely necessary: a few neurons often suffice, and in many chains disjoint subsets each flip the answer on their own, in an exploratory analysis most often when the model is less confident. Group effects transfer to rephrasings of the same chain but are much weaker on other tested chains. We study neuron-level causal interventions and alternative subsets; we do not recover the full connectivity of circuits.
\end{abstract}

\section*{1\quad Introduction}

A common form of evidence in mechanistic interpretability is a group intervention: a set of internal components is selected, removed or overwritten, and the model's output changes. Goodfire's recent essay on neuron interpretability gives a vivid example [1]. Qwen3-32B answers a riddle about the Pride Lands with ``Scar''; removing an 18-neuron ``Lion King'' group found in an attribution graph makes it answer ``Simba'' (P(Scar) = 0.24), whereas removing 18 random neurons from the same layers leaves P(Scar) at 0.91. The demonstration is a genuine causal experiment, and the random control is appropriate.

What such a demonstration does not show is the internal structure of the group. Would 17 neurons have been enough? 10? 4? Could a completely different set of four neurons produce the same change? Do the neurons act independently, so that their effects add up, or only in combination? And does the group do the same thing for a rephrased question, or for a different question? Without these answers it is hard to say how much explanatory weight ``these 18 neurons'' can carry. The public write-up does not report how the group size was chosen, what intervention was used to remove the neurons, or how subsets of the group behave; we treat these as open questions rather than as shortcomings of that work, whose broader claims do not rest on this one example.

Our central question is therefore: \textbf{when intervening on a group of neurons changes a model's answer, how much of that group is actually necessary?} We break it into three parts. \emph{Uniqueness:} can other, disjoint subsets produce the same change? \emph{Interaction:} are the neurons' effects additive, or do they depend on each other? \emph{Transfer:} does the same frozen group have the same effect on rephrasings of the question, and on other questions?

M\'eloux et al. have shown in small networks trained on toy tasks that mechanistic explanations need not be unique [16]. We ask the corresponding question for pre-trained language models, at the level at which group demonstrations are actually made: a specific, selected group of neurons. Because a group of 12 neurons has only 4,096 subsets, these questions can be answered exhaustively rather than by sampling. We do this in two studies and three model families:

\begin{itemize}
  \item \textbf{Study 1 (capital questions, Qwen3-0.6B).} Twelve neurons selected by zero-ablation effect pass the usual random-neuron control ($-$27\% margin vs. $\sim$0\%), their effects are almost perfectly additive and they transfer to rephrasings and new questions. Yet replacing them with values from \emph{other} questions has almost no effect. A group can look causal, additive and transferable without any evidence of question-specific content on the tested questions.
  \item \textbf{Study 2 (two-hop relation chains, Qwen3-1.7B, Llama-3.2-1B, Gemma-3-1B).} Patching 12 neurons with activations from the first-hop question reproduces the Scar $\rightarrow$ Simba shift in most chains of all three families, random neurons never do, and the frozen group has much smaller effects on other chains. Within the group, a few neurons usually suffice, and in many chains disjoint subsets each flip the answer.
  \item \textbf{A practical point for group demonstrations.} Exhaustive subset analysis is cheap at this scale, and a content-matched control (resampling or patching from other inputs) distinguishes generic from content-specific groups where a random-neuron control does not.
\end{itemize}

We study neuron-level causal interventions and alternative subsets; we do not recover the full connectivity of circuits. Our models and our neuron groups are our own; we did not test Goodfire's 18 neurons and do not claim to refute their result.

\section*{2\quad Related work}

\textbf{Circuits and group interventions.} Circuit analyses localize behaviours to sets of components and validate them by intervention, from factual recall [4] and indirect-object identification [5] to automated circuit discovery [6] and attribution graphs built with transcoders [2, 3]. Goodfire's essay [1] argues that many individual MLP neurons are already interpretable and illustrates it with neuron groups selected from such graphs.

\textbf{Redundancy and self-repair.} Wang et al. found backup heads that take over when primary heads are ablated [5], and McGrath et al. described the Hydra effect, in which downstream layers compensate for ablated ones [7]. These results suggest that a single ablation can under- or over-state a component's role; our exhaustive subset analysis quantifies this for neuron groups.

\textbf{Non-uniqueness of explanations.} The work closest to ours in question is that of M\'eloux et al. [16], who ask whether mechanistic interpretability is identifiable. On small networks trained on toy tasks, where every candidate explanation can be enumerated, they show that explanations are generally not unique: several circuits reproduce the same behaviour and one circuit admits several interpretations. Chen et al. found structurally different, faithful and sparse circuits for the same task in large language models, including an ultra-sparse one in which no single edge is indispensable [17]. Our results agree that sufficient neuron groups need not be unique. They differ in setting and scope: instead of enumerating explanations of a small network built for the purpose, we take pre-trained language models of three families on natural-language questions and exhaustively test every subset of the specific neuron group that a group demonstration would report. This turns the identifiability concern into a measurement on real models, and adds the observation that alternatives are common when the model is unsure of its answer and less common when it is confident.

\textbf{Choosing the intervention.} Zhang and Nanda showed that conclusions from activation patching depend on the corruption method and the metric [8], and attribution patching approximates patching effects with gradients [9]. Our Study 1 is a concrete instance of the first point: zero ablation and resampling disagree completely about the same neurons.

\textbf{Two-hop reasoning.} Language models often know both facts of a two-hop question yet fail to compose them (the compositionality gap [10]); whether and where they resolve the bridge entity internally has been studied with probing and patching [11, 12]. Closest to our Study 2, Arora et al. traced circuits in the MLP-neuron basis and found that small neuron sets encode intermediate steps of a multi-hop city--state--capital task and can be steered to change the output [18]; we ask how much of such a set is necessary and whether disjoint parts of it can each do the job. We use two-hop chains because they give a natural, content-specific alternative answer (the bridge).

\section*{3\quad Method}

\subsection*{3.1\quad Models, neurons and interventions}

We use the base (pre-trained) checkpoints of Qwen3-0.6B and Qwen3-1.7B [13], Llama-3.2-1B [14] and Gemma-3-1B [15]. The Llama and Gemma weights were obtained from the public unsloth mirrors of the official checkpoints. A \textbf{neuron} is one coordinate of the input to an MLP's down-projection, i.e. the gated MLP activation, at a given layer. All interventions act only at the last prompt token (the position from which the answer is generated) and leave all other positions untouched. We use four interventions on a neuron set S: \emph{zero} ablation; \emph{mean} ablation with the average activation over a separate set of prompts; \emph{resampling}, which writes the activation the neuron takes on a different prompt; and \emph{patching}, which writes the activation from a designated source prompt.

\subsection*{3.2\quad Metric}

For a prompt with correct answer token c and alternative token w, the margin is m = logit(c) $-$ logit(w). In Study 1, w is the other option letter; in Study 2, c is the first token of the final answer (``Scar'') and w the first token of the bridge entity (``Simba''). The effect of intervening on S is $\Delta$(S) = m(clean) $-$ m(S), reported relative to the clean margin m$_0$. $\Delta$/m$_0$ = 1 means the margin reaches zero; $\Delta$/m$_0$ > 1 means the bridge is now preferred (the answer \textbf{flips}). Exact ties, which are frequent in bfloat16, are not counted as flips. We also record whether the most likely next token becomes the bridge and generate greedily under the intervention.

\subsection*{3.3\quad Selecting 12 candidate neurons}

All neurons of the relevant layers are ranked by a first-order estimate of their effect, a$\cdot$$\partial$m/$\partial$a for zero ablation and (a\emph{target $-$ a}source)$\cdot$$\partial$m/$\partial$a for patching [9]. Real single-neuron interventions are then measured for every neuron of layers 14 and 21 in Study 1, and for the top 300 by estimated effect in Study 2. In both studies the 12 candidates are the 6 strongest by measured effect plus the next 6 by estimated effect, so that neurons that are weak alone are not excluded a priori. Candidates are selected on one prompt per question and frozen; rephrasings and other questions are never used for selection.

\subsection*{3.4\quad Exhaustive subset analysis}

For each frozen group we measure $\Delta$(S) for all $2^{12}$ = 4,096 subsets. From this table we derive: (i) \textbf{additivity}, the fit of $\Delta$(S) by the sum of single-neuron effects (R$^2$ and slope through the origin); (ii) \textbf{pairwise interactions} I(i, j) = $\Delta$(\{i, j\}) $-$ $\Delta$(\{i\}) $-$ $\Delta$(\{j\}); (iii) \textbf{minimal sets}: an effective subset ($\Delta$(S) above a threshold) none of whose proper subsets is effective; we use thresholds relative to the full group's effect (30/50/70\%) and, in Study 2, the flip threshold $\Delta$(S) > m$_0$; (iv) \textbf{alternatives}: pairs of minimal sets that share no neuron; and (v) \textbf{transfer}, by applying the same frozen group to other prompts.

\subsection*{3.5\quad Controls, noise and procedure}

Every result is compared with 12 random neurons matched in layer counts and receiving the same intervention (one random draw per question with a fixed seed; Study 1 also uses high-activation neurons with low estimated effect). Every batch of interventions contains an unmodified row, and $\Delta$ is computed against it. Study 1 runs in float32. Study 2 runs in bfloat16 because of GPU memory; there, we re-measured 30 subsets per chain with a different batch size and full (uncached) forward passes and report the maximum discrepancy as a noise floor (0.125--0.500 logits across models). Interaction thresholds are set to max(5\% of the full effect, three times the noise floor). Before each step, definitions and decision rules were written down; Appendix B lists every change made after seeing data. Confidence intervals are 95\% bootstrap intervals over questions (chains). For transfer, the resampled units are valid rephrasings or ordered source--target chain pairs, without adjustment for observations that share a chain.

\section*{4\quad Study 1: a causal-looking group without evidence of question-specific content}

\textbf{Setup.} Qwen3-0.6B answers 10 capital questions in a two-option format (``A. Paris / B. Stockholm''), with a three-example prompt. Candidates come from layers 14 and 21 and are selected by zero-ablation effect. A separate set of 10 questions is held out for testing.

\textbf{The group passes the usual checks.} Zero-ablating all 12 neurons lowers the margin by 0.27 [0.26, 0.30] ($\Delta$/m$_0$), while 12 random neurons give -0.00 [-0.00, 0.00]. The effects are almost perfectly additive (R$^2$ = 0.99 [0.99, 1.00], slope 1.00 [0.97, 1.02]), no pair of neurons interacts beyond the threshold (0 pairs in total), and many alternative subsets exist (15--306 minimal sets at 50\% of the full effect, disjoint pairs in 8/10 questions). The group transfers: across all 4,095 subsets, effects on rephrasings correlate with the original at Spearman 0.96--1.00. The same neuron, (14, 692), is the strongest single neuron for all 10 questions, and remains so when the options are swapped (10/10).

\textbf{But we find no evidence of question-specific content.} Replacing the 12 neurons with their values from other capital questions leaves the answer unchanged (same correct letter: 0.00 [-0.01, 0.01]; other correct letter: -0.01 [-0.02, -0.00]; mean of other capital questions: -0.00 [-0.01, 0.00]; Figure 1). The activations are nearly constant across questions: the median absolute deviation of a neuron from its mean activation on the other discovery questions is 0.12, against a median magnitude of 2.53. Zero ablation removes a nearly constant signal that the answer depends on; within the tested questions we see no sign that it removes information about Paris specifically.

\textbf{Replication on held-out questions.} Re-running the same selection procedure on the 10 test questions reproduces every pattern: additivity R$^2$ 0.989--0.996, 0 interacting pairs, and (14, 692) as the strongest neuron in 10/10 questions. The group frozen from the discovery questions lowers the margin on the 10 new questions by 0.24 [0.22, 0.26] (random group on the same prompts: -0.01 [-0.01, 0.00]).

\begin{figure}[htbp]\centering
\includegraphics[width=1.000\linewidth]{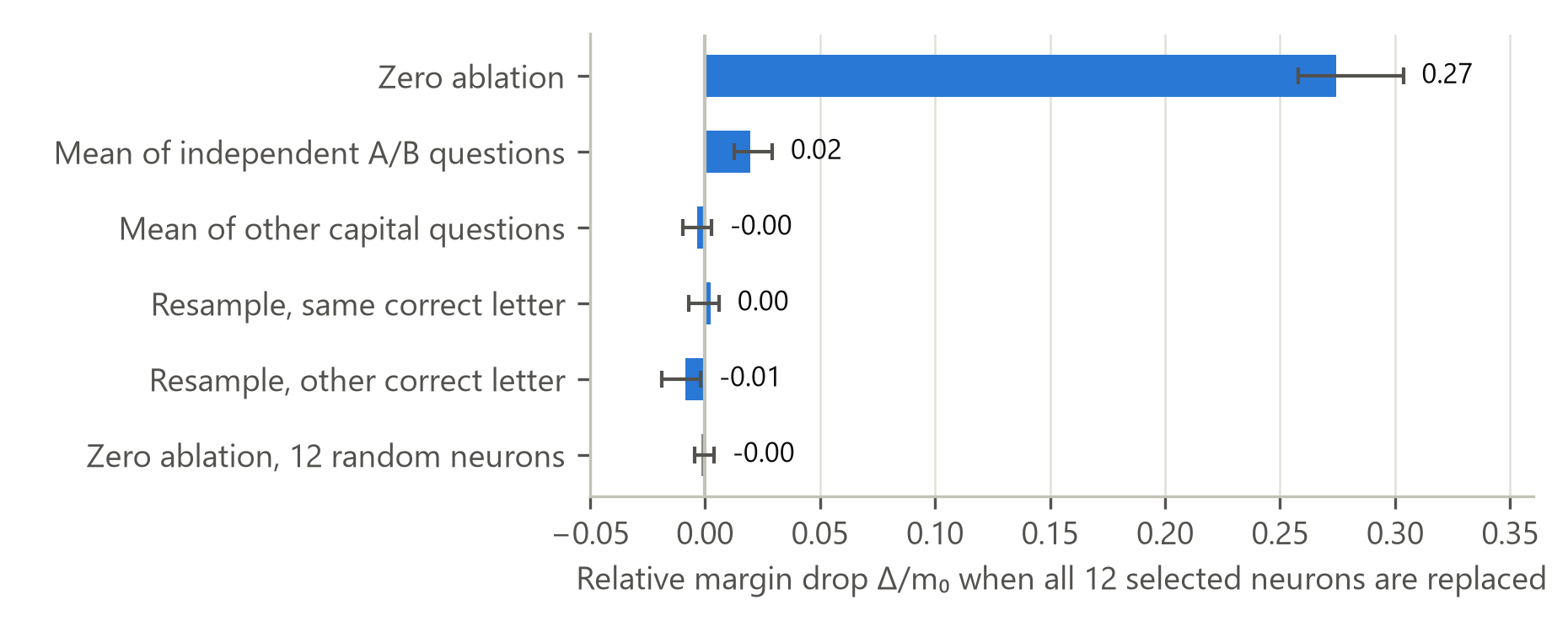}
\caption*{\textbf{Figure 1.} Study 1 (Qwen3-0.6B, capital questions). Relative margin drop when the 12 selected neurons are replaced, median and 95\% bootstrap CI over 10 questions. Zero ablation produces by far the largest margin reduction; mean ablation and resampling the same neurons from other questions have effects close to zero.}
\end{figure}

Study 1 shows that the standard evidence (a large effect, a null random control, additivity, transfer) is compatible with a group that is irrelevant to the question's content. This motivated Study 2, which uses a task where the alternative answer is itself content: the bridge entity.

\section*{5\quad Study 2: two-hop chains and the Scar $\rightarrow$ Simba shift}

\subsection*{5.1\quad Task and competence}

We wrote 20 two-hop chains in the spirit of the Lion King riddle: subject $\rightarrow$ bridge $\rightarrow$ answer (e.g. The Lion King $\rightarrow$ Simba $\rightarrow$ Scar; Eiffel Tower $\rightarrow$ France $\rightarrow$ Paris; Mona Lisa $\rightarrow$ Leonardo $\rightarrow$ Italy). Each chain has a first-hop question (answer: the bridge), a second-hop question (answer: the final answer), and three phrasings of the composed question (Appendix A). Before any neuron analysis we checked which models solve the chains (Table 1). All three know nearly every single hop but compose only about half of the chains, the compositionality gap [10]; their typical errors are to answer with the bridge itself (e.g. ``Denmark'' for the capital of Hamlet's country), the very shift that the Goodfire intervention induces. A chain is included when both hops and the first composed phrasing are answered correctly with m$_0$ $\geq$ 1.

\begin{table}[htbp]\centering\small
\caption*{\textbf{Table 1.} Competence on the 20 chains (greedy generation). Included = chains entering the neuron analysis.}
\begin{tabular}{>{\raggedright\arraybackslash}p{0.200\dimexpr\linewidth-14\tabcolsep\relax}>{\raggedright\arraybackslash}p{0.111\dimexpr\linewidth-14\tabcolsep\relax}>{\raggedright\arraybackslash}p{0.111\dimexpr\linewidth-14\tabcolsep\relax}>{\raggedright\arraybackslash}p{0.156\dimexpr\linewidth-14\tabcolsep\relax}>{\raggedright\arraybackslash}p{0.156\dimexpr\linewidth-14\tabcolsep\relax}>{\raggedright\arraybackslash}p{0.148\dimexpr\linewidth-14\tabcolsep\relax}>{\raggedright\arraybackslash}p{0.119\dimexpr\linewidth-14\tabcolsep\relax}}
\toprule
\textbf{Model} & \textbf{Hop 1} & \textbf{Hop 2} & \textbf{Composed (phr. 1)} & \textbf{Composed (phr. 2)} & \textbf{All four correct} & \textbf{Included} \\
\midrule
Qwen3-1.7B & 20/20 & 20/20 & 12/20 & 14/20 & 12/20 & 11 \\
Llama-3.2-1B & 19/20 & 20/20 & 12/20 & 11/20 & 10/20 & 11 \\
Gemma-3-1B & 20/20 & 20/20 & 11/20 & 11/20 & 9/20 & 10 \\
\bottomrule
\end{tabular}
\end{table}

\subsection*{5.2\quad Patching 12 neurons shifts the answer to the bridge}

For each chain we select a patching group (P): neurons whose activations, copied from the first-hop question (``Who is the main character of The Lion King?'', answer Simba), most reduce the margin of the composed question. For comparison we also select a zero-ablation group (Z). Patching the P group flips the answer in 10/11 chains (Qwen3-1.7B), 10/11 chains (Llama-3.2-1B), 8/10 chains (Gemma-3-1B); under greedy generation the model then starts its answer with the bridge in 9/11, 9/11, 8/10 chains. A generation counts as starting with the bridge only if its text begins with the bridge string (case-insensitive); ``Albert Einstein was born\ldots{}'' for the bridge Einstein is therefore not counted, which is why this count can be lower than the number of flips. The prefix count measures how the continuation begins, not whether it later mentions the correct answer as well. Twelve random neurons patched the same way never flip an answer (0/11, 0/11, 0/10). Table 2 shows the generations for Qwen3-1.7B.

\begin{table}[htbp]\centering\small
\caption*{\textbf{Table 2.} Greedy generations of Qwen3-1.7B under intervention on the full 12-neuron group (first line; ``\ldots{}'' marks truncation for display).}
\begin{tabular}{>{\raggedright\arraybackslash}p{0.134\dimexpr\linewidth-10\tabcolsep\relax}>{\raggedright\arraybackslash}p{0.220\dimexpr\linewidth-10\tabcolsep\relax}>{\raggedright\arraybackslash}p{0.207\dimexpr\linewidth-10\tabcolsep\relax}>{\raggedright\arraybackslash}p{0.232\dimexpr\linewidth-10\tabcolsep\relax}>{\raggedright\arraybackslash}p{0.207\dimexpr\linewidth-10\tabcolsep\relax}}
\toprule
\textbf{Chain} & \textbf{Answer $\rightarrow$ bridge} & \textbf{No intervention} & \textbf{P group patched} & \textbf{Z group zeroed} \\
\midrule
apple & Steve $\rightarrow$ Jobs & Steve Jobs & Steve Jobs & Jobs \\
beatles & England $\rightarrow$ Liverpool & England & Liverpool, England & Liverpool, England \\
eiffel & Paris $\rightarrow$ France & Paris & France & France \\
lisbon & Atlantic $\rightarrow$ Portugal & Atlantic Ocean & Portugal & The Atlantic Ocean \\
lk & Scar $\rightarrow$ Simba & Scar & Simba & Simba \\
madrid & Spanish $\rightarrow$ Spain & Spanish & Spain & Spanish \\
mona & Italy $\rightarrow$ Leonardo & Italy & Leonardo da Vinci was b\ldots{} & France \\
nairobi & Africa $\rightarrow$ Kenya & Africa & Kenya & Kenya \\
ninth & German $\rightarrow$ Beethoven & German & Beethoven & German \\
pyramids & Cairo $\rightarrow$ Egypt & Cairo & Egypt & Egypt \\
relativity & Germany $\rightarrow$ Einstein & Germany & Albert Einstein was bor\ldots{} & Albert Einstein was bor\ldots{} \\
\bottomrule
\end{tabular}
\end{table}

The two groups act differently (Figure 2). Patching both raises the bridge's logit (median change +4.1) and lowers the answer's logit (-3.7): it pushes the model \emph{towards} Simba as well as away from Scar. Zero ablation lowers the answer's logit more strongly (median -5.4) and raises the bridge less (+2.7; in Qwen3-1.7B the median bridge change is only +0.5). In the Z group, values taken from another chain are about as damaging as zeros (Qwen3-1.7B: zero 1.31 [0.88, 2.89], resample 1.37 [0.72, 2.70]): unlike in Study 1, these neurons are sensitive to replacement with values from another chain. The same holds for the P group: patching it with activations from the composed question of a \emph{different} chain also lowers the margin substantially ($\Delta$/m$_0$ 0.85 [0.02, 1.40], 0.92 [0.58, 1.25], 0.85 [0.28, 1.67]). Part of the patching effect is therefore a disruption of the answer that does not require the matching bridge, and we do not claim that the group encodes the bridge exclusively.

\begin{figure}[htbp]\centering
\includegraphics[width=0.752\linewidth]{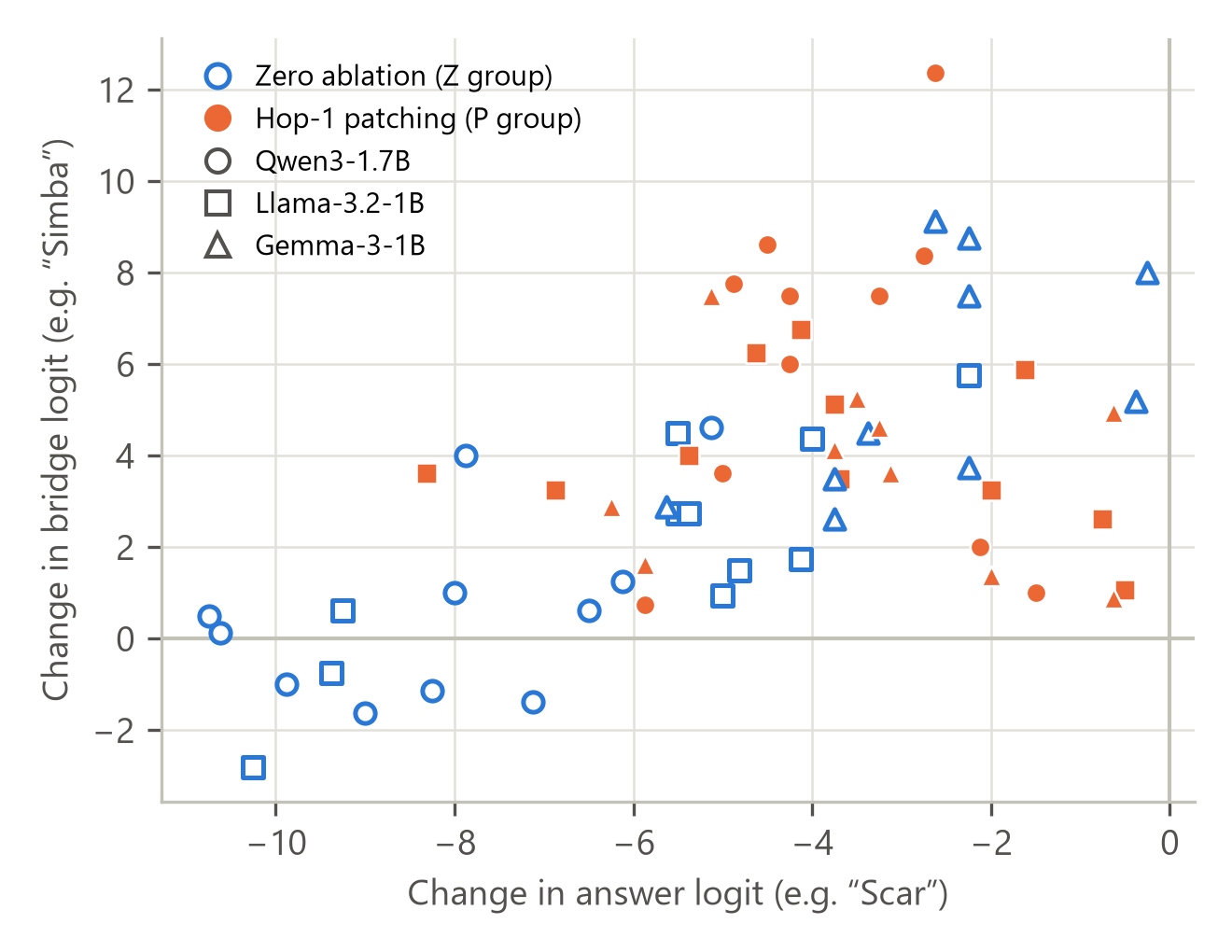}
\caption*{\textbf{Figure 2.} Change in the answer logit (x) and the bridge logit (y) when the full 12-neuron group is intervened on, one point per chain and model. Patching from the first-hop question (filled) raises the bridge; zero ablation (open) mainly lowers the answer.}
\end{figure}

\subsection*{5.3\quad How much of the group is necessary?}

Figure 3 addresses a question left open by the Goodfire example, how far the group can be reduced, for the Lion King chain: for every number of patched neurons it shows the distribution of effects over all subsets of that size. The full group is not necessary. The smallest answer-flipping subset has 4 neurons in Qwen3-1.7B, 8 in Llama-3.2-1B and 7 in Gemma-3-1B. The curves are smooth rather than step-like: effects accumulate neuron by neuron, and which neurons are chosen matters (the width of the 5--95\% band and the gap between the median and the strongest subset).

\begin{figure}[htbp]\centering
\includegraphics[width=1.000\linewidth]{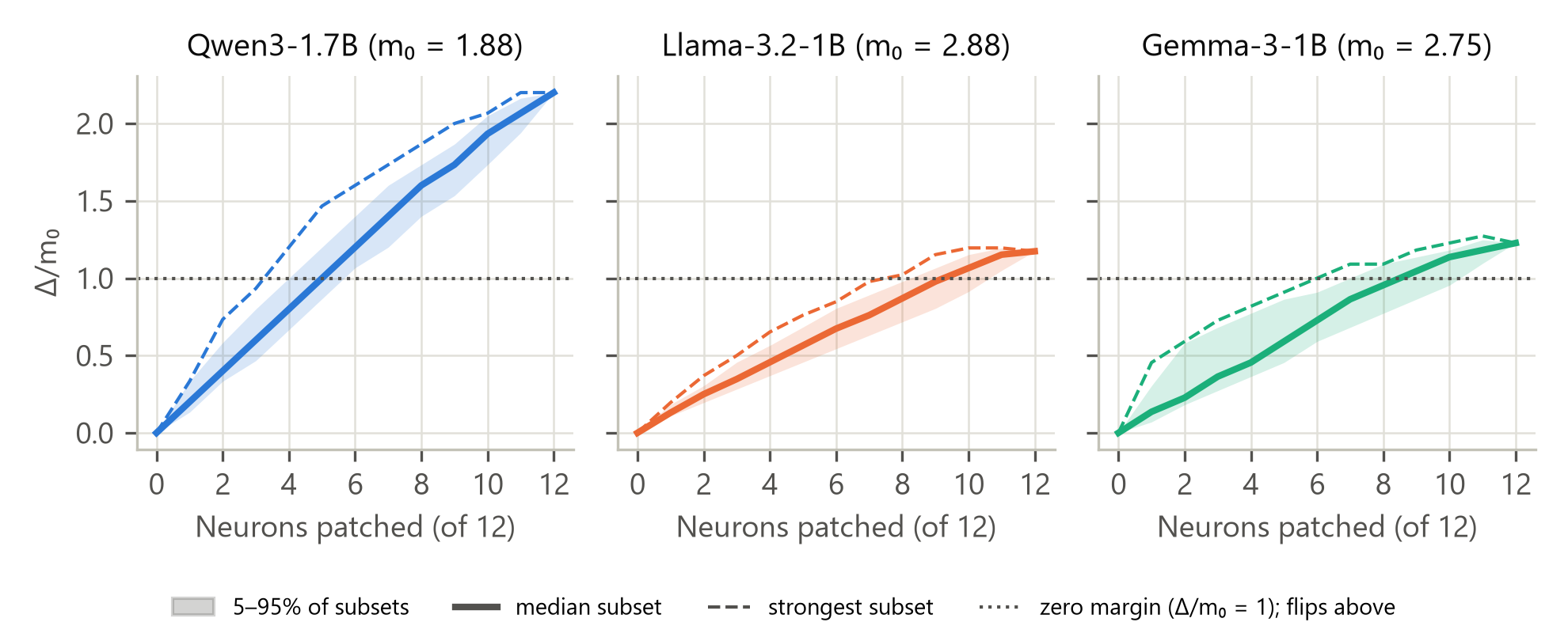}
\caption*{\textbf{Figure 3.} The Lion King chain. $\Delta$/m$_0$ against the number of patched neurons, over all subsets of each size (band: 5--95\%, solid: median, dashed: strongest subset). Above the dotted line the model prefers ``Simba''.}
\end{figure}

Across chains, the smallest flipping subset ranges over Qwen3-1.7B: 1, 2, 4, 4, 4, 4, 5, 5, 6, 6; Llama-3.2-1B: 1, 1, 1, 1, 2, 2, 2, 4, 5, 8; Gemma-3-1B: 1, 1, 1, 2, 2, 3, 4, 7 neurons (chains in which even the full group does not flip are omitted).

\subsection*{5.4\quad Uniqueness: disjoint subsets that flip the answer}

A key question is whether the same shift can be produced by \emph{different} neurons. We count pairs of minimal flipping subsets that share no neuron. Among the chains whose answer the full group flips, such pairs exist in 3/10 (Qwen3-1.7B), 7/10 (Llama-3.2-1B), 7/8 (Gemma-3-1B). In these chains the 12-neuron group is not a single indispensable unit: two non-overlapping parts of it each suffice. We note a caveat about our own examples: the four four-neuron subsets of Qwen3-1.7B that we checked by generation on the Lion King chain (each produces ``Simba'') all contain the same two neurons; disjoint alternatives are identified by the margin criterion and were not checked by generation.

Whether disjoint alternatives exist appears to depend on how confident the model is (Figure 4). Pooling the three models and counting only the 28 chains whose answer the full group flips, 13 of 15 chains with m$_0$ < 4 have disjoint flipping subsets, against 4 of 13 chains with m$_0$ $\geq$ 4. In the other 9 confident chains every pair of flipping subsets shares at least one neuron, so no disjoint alternative exists among the 12 selected neurons. The remaining 4 included chains are not flipped by the full group, and exhaustive enumeration finds no flipping subset in them either; all of them have m$_0$ $\geq$ 4.6. This is an exploratory observation: the cut-off of 4 was chosen after inspecting the Qwen3-1.7B results, and margins and chains are confounded.

\begin{figure}[htbp]\centering
\includegraphics[width=0.787\linewidth]{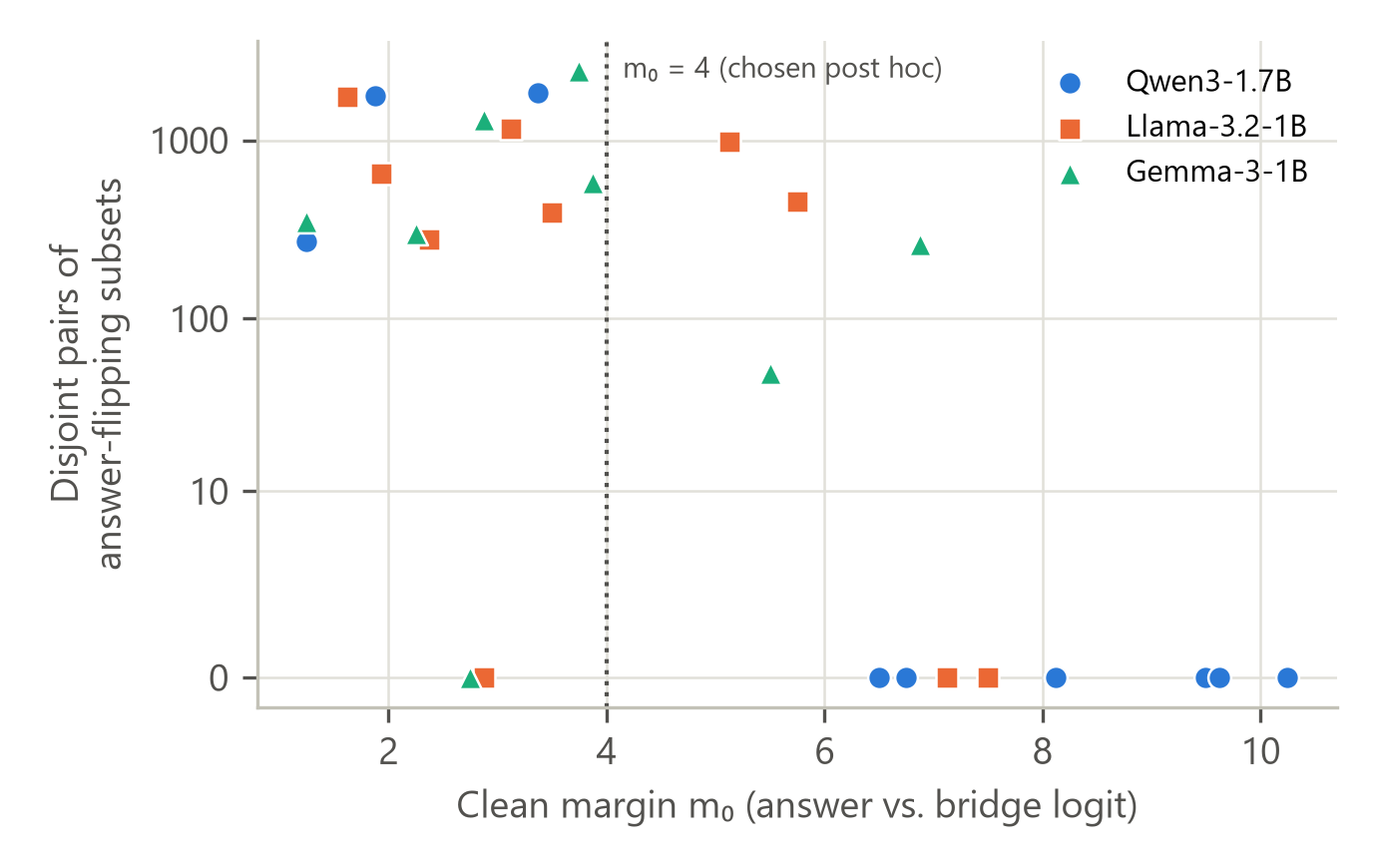}
\caption*{\textbf{Figure 4.} Number of disjoint pairs of minimal flipping subsets against the clean margin, one point per model and chain. Disjoint alternatives are more common below the margin cut-off than above it. Chains whose answer the full group does not flip are omitted. Exploratory.}
\end{figure}

\subsection*{5.5\quad Interaction}

Within the P groups, subset effects are well fit by a scaled additive model: R$^2$ of the additive fit is 0.90 [0.83, 0.92] (Qwen3-1.7B), 0.89 [0.70, 0.93] (Llama-3.2-1B), 0.84 [0.58, 0.92] (Gemma-3-1B), with slopes below one (0.78 [0.75, 0.90], 0.80 [0.68, 0.86], 0.83 [0.69, 0.90]), i.e. on average subsets do less than the sum of their neurons' single effects. Pairs of neurons whose joint effect departs from additivity by more than the threshold are rare (4 of 726 pairs (Qwen3-1.7B), 7 of 726 pairs (Llama-3.2-1B), 11 of 660 pairs (Gemma-3-1B)). We cannot exclude smaller interactions below the bfloat16 noise floor.

\subsection*{5.6\quad Transfer}

The ranking of subset effects carries over to other phrasings of the same chain (those with m$_0$ $\geq$ 1): across all 4,095 subsets, Spearman correlations with the original effects are 0.98 [0.97, 0.99] (Qwen3-1.7B), 0.98 [0.91, 0.98] (Llama-3.2-1B), 0.97 [0.90, 0.98] (Gemma-3-1B). Applied to the composed question of a \emph{different} chain (patching from that chain's own first-hop question), the same neurons have little effect: $\Delta$/m$_0$ 0.18 [0.07, 0.34] (Qwen3-1.7B), 0.09 [0.03, 0.26] (Llama-3.2-1B), 0.11 [0.04, 0.26] (Gemma-3-1B), against 1.52 [1.23, 2.20], 1.89 [1.35, 3.71], 1.82 [1.16, 3.17] on their own chain. Unlike in Study 1, the selected groups act much more strongly on their own chain than on the other tested chains.

\section*{6\quad The same pattern in three model families}

To compare models on identical questions we restrict to the 7 chains included for all three (apple, eiffel, lisbon, lk, madrid, nairobi, ninth). Table 3 and Figure 5 summarize the three answers. Every family shows the Scar $\rightarrow$ Simba shift under patching and none under random patching, strong transfer to rephrasings and weak transfer to other chains, and small sufficient subsets. The families differ in degree: disjoint alternatives are least common in Qwen3-1.7B, whose included chains also have the largest margins.

\begin{table}[htbp]\centering\small
\caption*{\textbf{Table 3.} The 7 chains shared by all three models. Medians with 95\% bootstrap CIs. Fractions use all 7 shared chains as denominator; subset sizes are listed only for chains the full group flips.}
\begin{tabular}{>{\raggedright\arraybackslash}p{0.327\dimexpr\linewidth-8\tabcolsep\relax}>{\raggedright\arraybackslash}p{0.224\dimexpr\linewidth-8\tabcolsep\relax}>{\raggedright\arraybackslash}p{0.224\dimexpr\linewidth-8\tabcolsep\relax}>{\raggedright\arraybackslash}p{0.224\dimexpr\linewidth-8\tabcolsep\relax}}
\toprule
\textbf{} & \textbf{Qwen3-1.7B} & \textbf{Llama-3.2-1B} & \textbf{Gemma-3-1B} \\
\midrule
Patching flips the answer & 6/7 & 6/7 & 6/7 \\
Generation starts with the bridge & 6/7 & 6/7 & 6/7 \\
Random neurons flip the answer & 0/7 & 0/7 & 0/7 \\
$\Delta$/m$_0$, full group & 1.52 [1.23, 1.62] & 1.83 [1.17, 2.21] & 1.80 [1.23, 3.17] \\
Smallest flipping subset (neurons) & 4, 4, 4, 4, 5, 6 & 1, 2, 2, 2, 4, 8 & 1, 1, 2, 3, 4, 7 \\
Chains with disjoint flipping subsets & 1/7 & 4/7 & 5/7 \\
Transfer: same chain, Spearman & 0.98 [0.94, 0.99] & 0.98 [0.90, 0.98] & 0.98 [0.93, 0.98] \\
Transfer: other chains, $\Delta$/m$_0$ & 0.01 [0.00, 0.04] & 0.02 [0.01, 0.25] & 0.03 [0.00, 0.10] \\
\bottomrule
\end{tabular}
\end{table}

\begin{figure}[htbp]\centering
\includegraphics[width=1.000\linewidth]{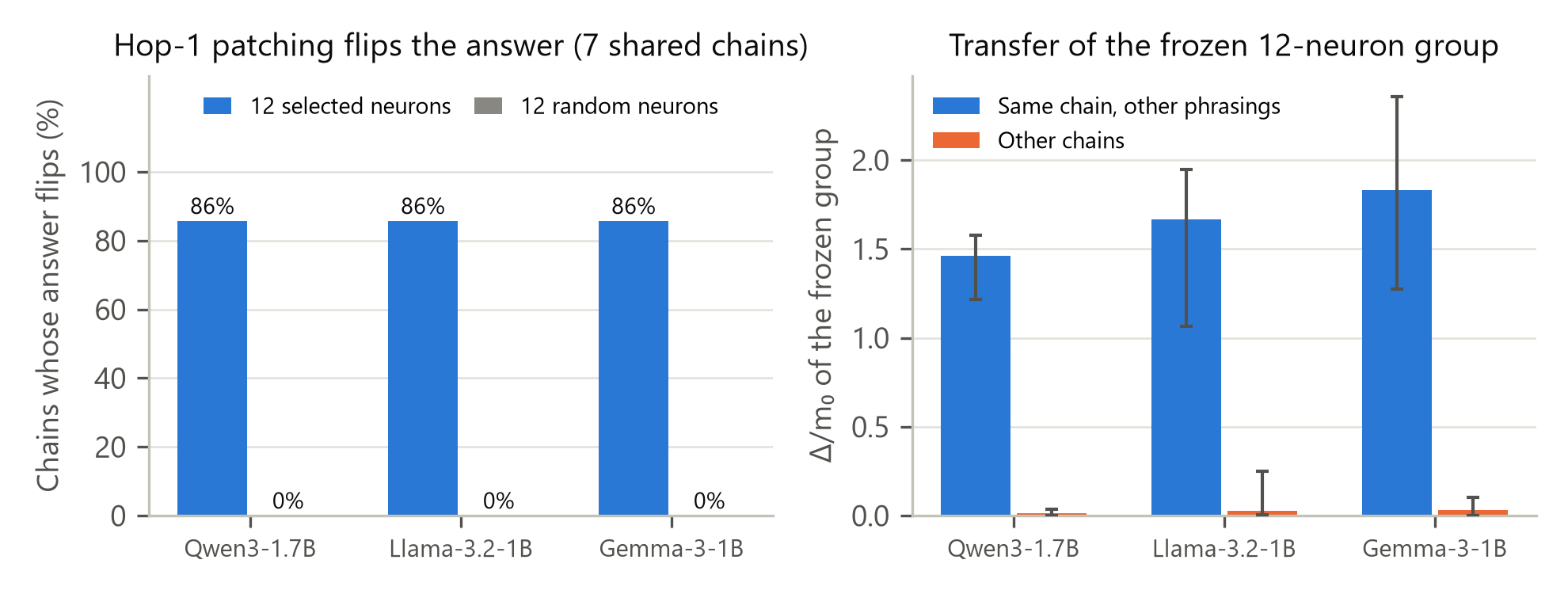}
\caption*{\textbf{Figure 5.} Left: share of the shared chains whose answer flips when 12 selected (blue) or 12 random (gray) neurons are patched. Right: effect of the frozen group on other phrasings of the same chain (blue) and on other chains (orange).}
\end{figure}

\section*{7\quad Discussion}

\textbf{How much of the group is necessary?} In our experiments, rarely all of it. A handful of neurons usually suffices to produce the shift, effects accumulate smoothly, and in many chains two disjoint parts of the group can each produce it. A statement such as ``removing these 18 neurons changes the answer'' is therefore an existence proof for \emph{a} sufficient intervention, not a description of the necessary mechanism. The flip of an answer also depends on how far the model was from flipping: when the margin is small, many small subsets cross the line and disjoint ones are common; when it is large, the flipping subsets usually overlap.

\textbf{Uniqueness is not a property of the group alone.} Whether a group looks unique depends on the threshold (half the effect, or a flip), on the margin, and on the 12 neurons that were selected. Minimal sets are minimal only relative to the candidate group; they are not the model's circuit.

\textbf{A random control is necessary but not sufficient.} Study 1 produced a group that beats random neurons, is additive, and transfers, yet showed no evidence of carrying anything specific to the tested questions. The control that exposed this was a content-matched one: resampling activations from other questions. In Study 2 the same checks behave differently: values from other chains disrupt the answer and the groups act much more weakly on other chains, so the groups are tied to their chain, although patching from another chain shows that part of the effect is not specific to the bridge. We suggest that group demonstrations report (i) a content-matched control in addition to random neurons, (ii) the size distribution of sufficient subsets, and (iii) the clean margin, since it shapes every necessity and uniqueness conclusion.

\textbf{Relation to the Lion King example.} We reproduced the \emph{kind} of effect reported by Goodfire in three smaller models from three families, using activation patching rather than removal, 12 rather than 18 neurons, and our own selection procedure. Our results do not say anything about their specific neurons; they show that the structural questions such an example leaves open have non-trivial and model-dependent answers.

\section*{8\quad Limitations}

\begin{itemize}
  \item \textbf{Scale.} Up to 11 chains per model and models of 0.6--1.7B parameters; results are pilot-scale estimates. Goodfire's example uses a 32B model, which did not fit in the GPU memory available to us.
  \item \textbf{Candidate groups.} All uniqueness and necessity statements are relative to 12 neurons chosen by single-neuron and first-order criteria, which favour neurons that matter individually and may miss purely synergistic ones.
  \item \textbf{Controls.} One layer-matched random group is drawn per question (fixed seed), not a distribution of random groups.
  \item \textbf{Single position.} Interventions act only on the last prompt token; computation at earlier positions is not tested.
  \item \textbf{Precision.} Study 2 runs in bfloat16 (noise floor up to 0.50 logits); small interactions may be invisible.
  \item \textbf{Exploratory analysis.} The margin cut-off in Section 5.4 was chosen after seeing data. Several definitions were revised between pilot runs (Appendix B).
  \item \textbf{Task format.} First-token margins between answer and bridge; multi-token answers are judged by the first token, generation checks the first line.
\end{itemize}

\section*{9\quad Conclusion}

A group intervention that changes a model's answer tells us that the group matters; it does not tell us how much of it is needed or whether other neurons could do the same. Measuring every subset of 12-neuron groups in four models, we find that the answer-changing effect is usually carried by a few neurons, that disjoint parts of a group can each produce it when the model is not confident, that such groups transfer to rephrasings but act much more weakly on other chains, and that a group can pass a random-neuron control without any evidence of carrying question-specific content. Neuron groups are useful handles on model behaviour, but their internal structure has to be measured, not assumed.

\section*{Code and data}

All code, the question sets, the saved results of every run reported here and the scripts that compute every number and figure in this paper are available at \url{https://github.com/omtekyav/causal-neuron-circuits} (MIT license).

\section*{Acknowledgment}

Generative AI tools were used for writing analysis code and for language editing. All experiments were run and all numbers were computed from saved result files by the author, who is responsible for the content.

\clearpage

\section*{Appendix A\quad The 20 two-hop chains}

\begin{table}[H]\centering\footnotesize
\caption*{\textbf{Table A1.} Chains, bridges and answers. Prompts use a three-example question--answer prefix.}
\begin{tabular}{>{\raggedright\arraybackslash}p{0.096\dimexpr\linewidth-10\tabcolsep\relax}>{\raggedright\arraybackslash}p{0.515\dimexpr\linewidth-10\tabcolsep\relax}>{\raggedright\arraybackslash}p{0.120\dimexpr\linewidth-10\tabcolsep\relax}>{\raggedright\arraybackslash}p{0.114\dimexpr\linewidth-10\tabcolsep\relax}>{\raggedright\arraybackslash}p{0.156\dimexpr\linewidth-10\tabcolsep\relax}}
\toprule
\textbf{Chain} & \textbf{Composed question (phrasing 1)} & \textbf{Bridge} & \textbf{Answer} & \textbf{Included in} \\
\midrule
hp & What is the real name of the wizard who killed Harry Potter's parents? & Voldemort & Tom & --- \\
lk & Who killed the father of the main character of The Lion King? & Simba & Scar & Qwen3, Llama, Gemma \\
mona & In which country was the painter of the Mona Lisa born? & Leonardo & Italy & Qwen3 \\
eiffel & What is the capital of the country where the Eiffel Tower is located? & France & Paris & Qwen3, Llama, Gemma \\
madrid & What language is spoken in the country whose capital is Madrid? & Spain & Spanish & Qwen3, Llama, Gemma \\
tokyo & What is the currency of the country where Tokyo is located? & Japan & yen & Llama \\
nairobi & On which continent is the country whose capital is Nairobi? & Kenya & Africa & Qwen3, Llama, Gemma \\
alex & Who was the teacher of Alexander the Great's teacher? & Aristotle & Plato & --- \\
romeo & In which town was the author of Romeo and Juliet born? & Shakespeare & Stratford & --- \\
lisbon & Which ocean borders the country whose capital is Lisbon? & Portugal & Atlantic & Qwen3, Llama, Gemma \\
titanic & Who directed the movie in which Jack and Rose meet on a sinking ship? & Titanic & James & Llama \\
ninth & What was the native language of the composer of the Ninth Symphony with the Ode to Joy? & Beethoven & German & Qwen3, Llama, Gemma \\
beatles & In which country is the city where the Beatles were formed? & Liverpool & England & Qwen3 \\
relativity & In which country was the scientist who developed the theory of relativity born? & Einstein & Germany & Qwen3, Llama \\
hamlet & What is the capital of the country where Shakespeare's Hamlet is set? & Denmark & Copenhagen & --- \\
pyramids & What is the capital of the country where the Pyramids of Giza are? & Egypt & Cairo & Qwen3, Gemma \\
kangaroo & What is the capital of the country famous for kangaroos? & Australia & Canberra & --- \\
sherlock & In which country was the creator of Sherlock Holmes born? & Doyle & Scotland & Gemma \\
apple & What was the first name of the person who co-founded Apple with Steve Wozniak and later led the company? & Jobs & Steve & Qwen3, Llama, Gemma \\
everest & What is the capital of the country on the south side of Mount Everest? & Nepal & Kathmandu & Llama, Gemma \\
\bottomrule
\end{tabular}
\end{table}

\section*{Appendix B\quad Pre-specified decisions and changes}

\begin{itemize}
  \item Pilot (Study 1): the effect threshold was first fixed at 50\% of the clean margin. No subset reached it (all 4,096 measured), so for the uniqueness analysis it was replaced, before further data were collected, by thresholds relative to the full group's effect; the 50\%-of-margin result is reported unchanged.
  \item The original minimality check only removed one neuron at a time; it was corrected to require that no proper subset be effective (relevant when effects are not monotone). Results reported here use the corrected definition.
  \item The first pilot computed mean ablation from the same questions (including the target); it was replaced by an independent reference set, and resampling from other questions was added.
  \item Study 1 was re-run in float32 after bfloat16 logits proved too coarse for single-neuron effects; Study 2 uses bfloat16 with a measured noise floor.
  \item Study 2 definitions, inclusion rules and success criterion were written before the first run; the same frozen protocol was applied unchanged to Llama-3.2-1B and Gemma-3-1B. The margin cut-off (m$_0$ = 4) in Section 5.4 is post hoc.
  \item After review, the counts of disjoint alternatives in Section 5.4 and Figure 4 were restricted to chains whose answer the full group flips (exhaustive enumeration finds no flipping subset in the four excluded chains), and the subset flip threshold was changed from $\Delta$(S) $\geq$ m$_0$ to $\Delta$(S) > m$_0$ so that exact ties are not counted, matching the full-group definition. This changes the Qwen3-1.7B disjoint-alternative count from 4/10 to 3/10 and three minimal subset sizes.
\end{itemize}

\end{document}